%% file: iclr2027_conference.tex
\documentclass{article} 
\usepackage{iclr2027_conference,times}

\input{math_commands.tex}

\usepackage{hyperref}
\usepackage{url}
\usepackage{amsmath}
\usepackage{amssymb}
\usepackage{mathtools}
\usepackage{amsthm}
\usepackage{listings}
\usepackage{fvextra}
\usepackage{multicol}
\usepackage{multirow}
\usepackage{xspace}
\usepackage{graphicx}
\usepackage{tcolorbox}
\usepackage[frozencache,cachedir=minted-cache]{minted}
\usepackage{xcolor}
\usepackage{bm}
\usepackage{enumitem}
\usepackage{adjustbox}
\usepackage{pgfplots}  
\usepackage{algorithm}
\usepackage[noend]{algpseudocode}
\usepackage{booktabs}
\usepackage{wrapfig}
\usepackage{etoc}  

\usepackage[utf8]{inputenc}
\DeclareUnicodeCharacter{2115}{\ensuremath{\mathbb{N}}}
\usepackage{newunicodechar}
\newunicodechar{≡}{\ensuremath{\equiv}}
\newunicodechar{｜}{\textbar}
\newunicodechar{₀}{\ensuremath{_0}}
\newunicodechar{₁}{\ensuremath{_1}}
\newunicodechar{₂}{\ensuremath{_2}}
\newunicodechar{₃}{\ensuremath{_3}}
\newunicodechar{₄}{\ensuremath{_4}}
\newunicodechar{₅}{\ensuremath{_5}}
\newunicodechar{₆}{\ensuremath{_6}}
\newunicodechar{₇}{\ensuremath{_7}}
\newunicodechar{₈}{\ensuremath{_8}}
\newunicodechar{₉}{\ensuremath{_9}}
\newunicodechar{∨}{\ensuremath{\vee}}
\newunicodechar{ℚ}{\ensuremath{\mathbb{Q}}}
\newunicodechar{≠}{\ensuremath{\neq}}
\newunicodechar{ℤ}{\ensuremath{\mathbb{Z}}}
\newunicodechar{✝}{\ensuremath{\dagger}}
\newunicodechar{∣}{\ensuremath{\mid}}
\newunicodechar{ℝ}{\ensuremath{\mathbb{R}}}
\newunicodechar{∧}{\ensuremath{\land}}
\newunicodechar{≤}{\ensuremath{\leq}}
\newunicodechar{⊢}{\ensuremath{\vdash}}
\newunicodechar{ℂ}{\ensuremath{\mathbb{C}}}
\newunicodechar{▁}{\textunderscore}  
\newunicodechar{↔}{$\leftrightarrow$}
\newunicodechar{ˣ}{$^x$}
\newunicodechar{↦}{$\mapsto$}
\newunicodechar{∃}{$\exists$}
\newunicodechar{∈}{$\in$}
\newunicodechar{≥}{$\geq$}

\def\btau{\bm{\tau}}

\definecolor{verylightgray}{gray}{0.95}  
\newcommand{\method}{\textbf{{Lean4AgenticRL}}\xspace}

\title{Lean4AgenticRL: Advancing Agentic RL with Verifiable Multi-level Process Rewards}

\author{
\hspace*{-0.1in}%
Ruida Wang$^{1,2,*}$
\hspace{-0.2em}
Gerald Friedland$^{1}$
\hspace{-0.2em}
Tong Zhang$^{2}$
\hspace{-0.2em}
Chris Kong$^{1,\dagger}$
\\
$^{1}$Amazon Web Services
\qquad
$^{2}$University of Illinois Urbana-Champaign
}

\iclrfinalcopy 
\begin{document}

\maketitle

\begingroup
\renewcommand{\thefootnote}{\fnsymbol{footnote}}
\footnotetext[1]{Work done during an internship at Amazon Web Services.}`
\footnotetext[2]{Correspondence to: Chris Kong at \texttt{luyankon@amazon.com}.}
\endgroup

\etocdepthtag.toc{mainbody}
\input{sec/0_abstract}
\input{sec/1_introduction}
\input{sec/3_methodology}
\input{sec/4_experiment}
\input{sec/2_related_work}
\input{sec/5_conclusion}
\input{sec/6_others}

\newpage
\bibliography{iclr2027_conference}
\bibliographystyle{iclr2027_conference}

\input{sec/X_Appendix}

\end{document}

%% file: math_commands.tex
\usepackage{amsmath,amsfonts,bm}

\def\eqref#1{equation~\ref{#1}}

\def\1{\bm{1}}

\DeclareMathAlphabet{\mathsfit}{\encodingdefault}{\sfdefault}{m}{sl}
\SetMathAlphabet{\mathsfit}{bold}{\encodingdefault}{\sfdefault}{bx}{n}



%% file: sec/0_abstract.tex
\begin{abstract}
Enhancing the agentic capabilities of Large Language Models (LLMs) to tackle complex software engineering (SWE) and general terminal-usage problems is a pressing challenge. Current state-of-the-art LLMs often go through large-scale agentic Reinforcement Learning (RL). However, they mostly rely on high-level outcome rewards~\citep{zeng2026glm}, with minimal consideration of low-level toolcall and mid-level workflow-stage rewards, which makes the training less effective. To overcome this limitation, we propose \method, which, to the best of our knowledge, is the first framework that grounds multi-level process rewards in the unified Lean4 formalization schema for open-ended agentic RL. \method first builds a \textit{problem-specific workflow} system that balances the agent's own capability and controllability during execution, laying the foundation for further verification. Subsequently, our workflow formalization and multi-level reward annotation methods enable agent behavior verification within a general harness system. It provides multi-level rewards to support more effective agentic RL with the workflow-stage-normalized GRPO algorithm. Using \method, we train \textbf{LAR-9B/27B}, a family of models with enhanced agentic capability. Extensive experiments indicate that our model outperforms the leading model of the similar size. \textbf{LAR-27B} achieves \textbf{77.3\%} on TerminalBench-2.1. On average, our models outperform the base model by \textbf{9.89\%} and surpass the second-best model of comparable size we know by \textbf{5.07\%}. These results indicate the potential of integrating formal methods in enhancing open-ended agentic RL.
\end{abstract}

%% file: sec/1_introduction.tex
\section{Introduction}\label{sec:intro}

The ability of Large Language Models (LLMs) to perform long-running agentic tasks has been regarded as a key capability by the computer science community~\citep{wang2026lean4agent}. This capability is typically evaluated by solving complex software engineering (SWE)~\citep{jimenez2024swebench} and terminal-usage~\citep{tbench_2025} tasks within advanced harness frameworks~\citep{anthropic2026claudecode, openai2025codex, merrill2025terminus}. Given the significant practical value~\citep{openai2026gpt56}, training advanced agentic LLMs has become an important research direction.
Agentic capability in LLMs refers to the model's ability to actively interact with an environment to explore, understand, and resolve tasks. Previous works have developed multiple approaches to strengthen this ability, including simple prompting~\citep{yao2022react, madaan2023self} and building supervised fine-tuning datasets~\citep{pan2024training, badertdinov2026swe}.
Building on the success of scaling Reinforcement Learning with Verifiable Rewards (RLVR)~\citep{guo2025deepseek, team2025kimiK1.5} and multi-turn agentic RL methods~\citep{pan2024training, wei2026swe, wang2026openclaw}, the community has driven significant progress in long-horizon agent tasks such as SWE and terminal-usage.

\begin{figure*}[t]
    \vspace{-0.3in}
    \centering
    \includegraphics[width=0.95\columnwidth]{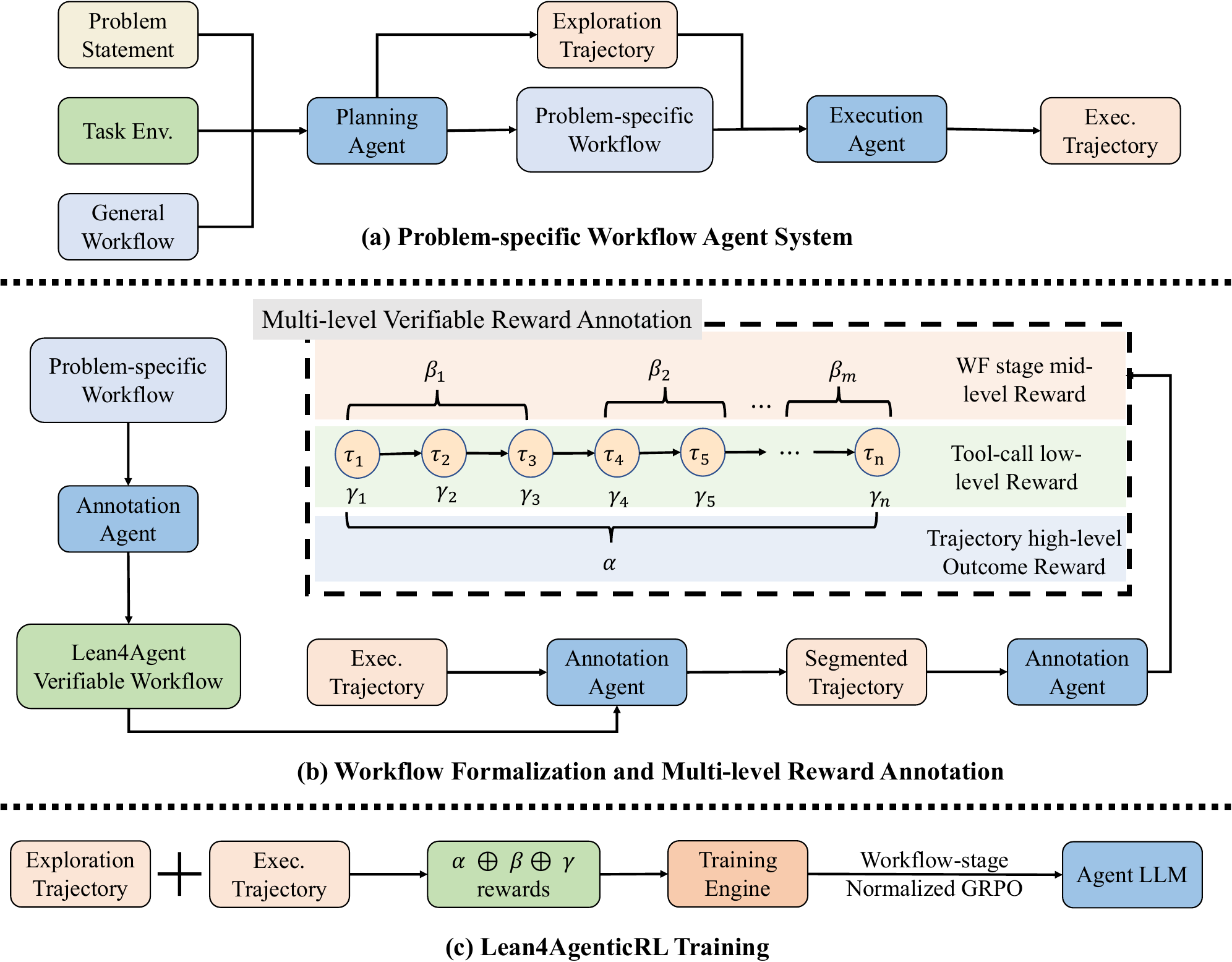}
    \vspace{-0.15in}
    \caption{
    \textbf{Overview of \method framework:} The \method framework contains three major steps: 
    (a) \textit{Problem-specific workflow agent system}: A plan-then-execute process that first explores and produces the problem-specific workflow (PSWF). Then execute the PSWF to solve the task.
    (b) \textit{Workflow Formalization and Multi-level Reward Annotation}: An annotation process that formalizes the PSWF into our generalized Lean4Agent framework. Then segment the agent trajectory into workflow stages, and assign mid-level workflow-stage rewards and low-level toolcall rewards.
    (c) \textit{\method training}: We apply the workflow-stage-normalized GRPO algorithms to train the LLM agent with multi-level rewards.
    }
    \label{fig:main}
    \vspace{-0.25in}
\end{figure*}

Despite significant progress in large-scale agentic RL, reliably assigning process rewards in long agent traces remains an open problem.
In particular, current agentic RL methods~\citep{zeng2026glm, xu2026deepseek, team2026kimi3} primarily rely on high-level outcome rewards, building on their success in single-turn RLVR. However, these signals do not directly distinguish useful intermediate actions from incorrect or redundant steps within successful trajectories, limiting the effectiveness of training~\citep{liu2026agentic}. This has motivated the use of process rewards to supervise the agent's intermediate steps~\citep{wang2026openclaw, xi2026agentprm}. However, existing approaches often focus on low-level toolcall results, which may overlook broader workflow progress and degrade performance when a single toolcall cannot directly connect to the task objectives~\citep{modecrua2026multi}. Overall, these limitations highlight the need to unify the high-level outcome rewards, mid-level workflow-stage rewards, and low-level toolcall rewards in a verifiable way to enable more effective agentic RL.

To address the above limitations, we propose \method, a unified framework for deriving multi-level rewards within a uniform, Lean4Agent verifiable, open-ended agentic RL setting. The overview of \method is presented in Figure~\ref{fig:main}. In \method, we first use \textit{problem-specific workflow} (PSWF) agent system that balances the agent's inherent capabilities with execution controls through a plan-then-execute paradigm. It further provides the basis for verifying agent behaviors. 
Subsequently, we extend the Lean4Agent~\citep{wang2026lean4agent} verification framework from fixed workflows to general harnesses. Then, we formalize the PSWF and segment the agent trajectories into formal workflow stages and annotate the trajectories with high-level outcome rewards ($\alpha$), mid-level workflow-stage rewards ($\beta$), and low-level toolcall rewards ($\gamma$). These reward signals synergistically support learning from both final outcomes and intermediate behaviors through our workflow-stage-normalized GRPO algorithm, enabling more effective agentic RL.

We summarize our contributions as follows:
(1) We propose \method, which, to the best of our knowledge, is the first framework that derives multi-level process rewards within Lean, a uniform dependent-type verifiable language, for agentic RL in open-ended domains that enables more effective agentic RL.
(2) We develop \textit{problem-specific workflow} and generalize the current agent verification framework from fixed workflow to general harness systems. They greatly enlarge the applicability of the current uniform formal agent verification system.
(3) We train \textbf{LAR-9B/27B} through \method. Extensive experiments indicate that our model outperforms the leading model of comparable size on SWE and general terminal-usage tasks. \textbf{LAR-27B} achieves \textbf{77.3\%} on TerminalBench-2.1~\citep{tbench_2025} and, across the three benchmarks we evaluated, our model outperforms their base model by \textbf{9.89\%} and the second-best models of comparable size by \textbf{5.07\%} on average.
The performance highlights the effectiveness and generality of \method.

Furthermore, \method offers a new view of agentic RL methods grounded in formal verification and provides a unified framework for future studies to explore applications of formal methods within the modern RL paradigm. To facilitate the advancement of the fields, our code and model have been open-sourced in \href{https://github.com/RickySkywalker/Lean4AgenticRL}{https://github.com/RickySkywalker/Lean4AgenticRL}

%% file: sec/3_methodology.tex
\section{Methodology}\label{sec:meth}

We detail the formulation of the \method framework in this section. The goal of \method is to derive multi-level rewards in a uniform Lean schema to achieve more effective agentic RL in open-ended domains.
Section~\ref{meth:prelim} introduces the key preliminaries, Section~\ref{meth:prob} details the formulation of \textit{Problem-specific workflow} (PSWF) system, Section~\ref{meth:formal} introduces the workflow formalization and reward annotation method, and Section~\ref{meth:train} shows how we train the model using the workflow-stage-normalized GRPO method. 

\subsection{Preliminaries}\label{meth:prelim}
In this section, we introduce the central preliminary concept for \method. Additionally, we present the definitions of the terms we used in the paper in Appendix~\ref{appendix:term}.

\noindent \textbf{LLM Agent} Following ReAct~\citep{yao2022react}, we define an LLM agent as an LLM-based system that cycles between reasoning with tool-based actions. The reasoning enables the agent to formulate, track, and revise its plan, while actions allow it to interact with environments to execute external tools or read information sources. In our paper, we use Claude-code~\citep{anthropic2026claudecode} or Terminus-2~\citep{merrill2025terminus} as our agent harness.

\noindent \textbf{Lean4Agent Verification System} Following Lean4Agent~\citep{wang2026lean4agent}, we present the workflow for the LLM agent as a heterogeneous directed graph $\mathcal{G} = \langle V, E\rangle$, where $V:= \{v_i\}_{i = 1}^m$ contains the \textit{workflow stages} and $E \subseteq V \times V$ specifies the allowed transitions between stages. The Lean4Agent system implements a set of pre-conditions $\rho_i^{(pre)}$ and post-conditions $\rho_i^{(post)}$ with the \textit{predicates}, which describe atomic properties of the agent behaviors. When executing the workflow, we obtain trajectory $\bm{\tau}$, and each stage $i \in [1, m]$ has its trajectory $\bm{\tau}_i = \{\tau_{i, j}\}$. The trajectory-level verification evaluates whether its entry and exit states satisfy $\rho_i^{(pre)}$ and $\rho_i^{(post)}$, respectively. These predicate-based specifications, combined with a dependent-type formal foundation, provide the verification of workflow-level coherence and assess completion of the specified stages' goals. Following Lean4Agent, in this work, we define a stage as verified when its Lean4Agent workflow specification's post-condition holds over after the stage's execution.

\subsection{Problem-specific Workflow Agent System}\label{meth:prob}

This section introduces the \textit{problem-specific workflow} (PSWF) system. 
It aims to balance the controllability of fixed-workflow execution against the flexibility of free-form, ReAct-style execution.
It also lays the foundation for our further verification process. In particular, we achieve this by a plan-then-execute paradigm.

\subsubsection{Planning Phase}\label{prob:exp}

In this phase, we apply the LLM as the planning agent within the agent harness to explore the task environment and formulate a PSWF. Formally, given the problem statement $\mathcal{S}$, its environment $\mathcal{E}$ (commonly a container), and a general workflow $\mathcal{W}$, we obtain the PSWF $\mathcal{W}_{\mathcal{S}}$ as follows:
\[
\mathcal{W_S}, \btau^{(p)} = \text{H}(LLM(\mathcal{E}, \mathcal{S}, \mathcal{W}))
\]
where H$(\cdot)$ denotes the agent harness and $\btau^{(p)}$ denotes the trajectory for planning phase. The $\mathcal{W_S}$ is a natural-language plan that specializes in stage-wise guidance in $\mathcal{W}$ based on the findings from its exploration. For example, in SWE tasks, it contains detailed information such as relevant source files, suspected causes of the issue, and proposed repair and verification steps, providing concrete instructions for subsequent execution. The PSWF aims to control the agent behavior without limiting its flexibility.

\subsubsection{Execution Phase}\label{prob:exe}

In this phase, we execute the PSWF $\mathcal{W}_{\mathcal{S}}$ in its task environment $\mathcal{E}$ by instructing the LLM agent. To avoid repeated exploration, we also include the planning-phase trajectory $\btau^{(p)}$ in the context. Formally, we have:
\[
\mathcal{E}', \bm{\tau}^{(e)} = \text{H}(LLM(\mathcal{E}, \mathcal{W_S}, \bm{\tau}^{(p)}))
\]
where $\mathcal{E}'$ is the updated environment to resolve the problem and $\btau^{(e)} = \{\tau_i\}_{i = 1}^n$ records the execution steps. Alongside the planning phase, we complete the rollout of our agentic RL and lay the foundation for subsequent multi-reward assignment based on the agent verification system.

\subsection{Workflow Formalization and Multi-level Reward Annotation}\label{meth:formal}

In this section, we introduce methods for adapting a formal verification system beyond a fixed workflow schema via the PSWF formalization, and detail methods for annotating multi-level rewards based on the formalized workflow and execution trajectory.

\subsubsection{Generalizing Agent Verification System with PSWF}\label{formal:lean4agent}

The current agent verification framework that can uniformly process both workflow and trajectory levels, such as Lean4Agent~\citep{wang2026lean4agent}, is designed for predefined, fixed workflows. However, most modern agent harnesses are open-ended without such fixed workflows. Thus, to make \method applicable to a broader domain, we propose methods to generalize the verification framework with PSWF.
Although $\mathcal{W}_\mathcal{S}$ provides stage-wise guidance for $\btau^{(e)}$, the correspondence between agent steps and workflow stages must be determined from annotation. To do so, we need to first formalize the natural language $\mathcal{W}_\mathcal{S}$ to its formal counterpart
\[
\mathcal{F_W^{(S)}} = \{(v_i, \rho_{i}^{(pre)}, \rho_i^{(post)})\}_{i = 1}^{m_{\mathcal{S}}} = \text{H}(LLM(\mathcal{W_S}, \mathcal{W}))
\]
where $\mathcal{F_W^{(S)}}$ is the formalized workflow, $m_\mathcal{S} = |V|$ is the number of stages in $\mathcal{W_S}$. Each stage is also aligned with the general workflow $\mathcal{W}$, ensuring the stages remain comparable across rollouts. Subsequently, guided by $\mathcal{F_W^{(S)}}$, we sequentially assign each step of $\btau^{(e)}$ to the workflow stage as follows:
\[
j_i = LLM(\tau_{i}, \tau_{i + 1}, \tau_{i - 1}, \mathcal{F_W^{(S)}}) \in \{j_{i - 1}\} \cup \text{Connect}(v_{j - 1})
\]
where $j_i$ is the index of the stage to which $\tau_i$ is assigned, and Connect$(v_{j - 1})$ is the set of stages reachable from $v_j$ through an edge in $E$, including the loop-back edges. Here, each agent step either remains in the current stage or transitions to one of the connected stages. Since every decision depends only on the local window, not the full rollout, the segmentation scales to ultra-long trajectories. For loops and branches, each revisit of a stage is treated as a separate instance with its own index. Consequently, each stage $i$ obtains its step set $\{\tau_{i, j}\}_{j = 1}^{n_i} \subseteq \btau^{(e)}$, where $n_i$ denotes the number of steps assigned to it. Together, the formalized workflow and step-to-stage alignment allow the formal agent verification to move beyond fixed workflow to general agent harnesses.

\subsubsection{Workflow-Stage Mid-level Reward Annotation}\label{formal:reward}

After the execution trajectory is segmented into workflow stages. Following Lean4Agent~\citep{wang2026lean4agent}, we annotate the workflow-stage mid-level reward $\beta$ by verifying that each stage accomplishes its specified goals, represented as functions of predicates. This reward serves as the middle ground between the low-level toolcall rewards and high-level outcome rewards. The former is prone to overlooking broader workflow progress, and the latter cannot verify the correctness of intermediate actions. 
Specifically, for each stage $i$, we examine whether the steps $\{\tau_{i, j}\}_{j = 1}^{n_i}$ entered the specified pre-condition $\rho^{(pre)}_i$ and established the desired post-condition $\rho^{(post)}_i$. 
Directly decidable predicates are verified through Lean or Python code, while those requiring semantic interpretation are assessed by an LLM judge. The boundary for the formal components in our system is discussed in Appendix~\ref{appendix:formal_boundary}.

Formally, the workflow $\beta$ reward can be defined as: If $\rho^{(post)}$'s additional predicate requirements are satisfied upon the exit of stage $i$, then $\beta=1$, otherwise, $\beta = -1$. The detailed reward annotation schema is provided in Appendix~\ref{reward:beta}.

\subsubsection{Toolcall Low-level Reward Annotation}\label{formal:tool}

Since part of the predicate judgment in $\beta$ rewards relies on LLM-as-judge verification, the policy may learn to hack the reward through ritualistic behaviors rather than resolving the task. For example, to satisfy the predicate that requires repeated fix-and-retry attempts, the agent can simply rerun the same testing bash command several times and thereby mislead the judge.

To address this issue, we further extend the formal predicate system from the workflow level to a finer-grained toolcall level. Specifically, we implement a set of symbolic predicates, including \texttt{consecutive\_repetition}, \texttt{noop\_commands}, \texttt{const\_shell}, $\cdots$, and evaluate them using Lean and Python checkers to target these common reward-hacking patterns. The full list of penalized behaviors is presented in Appendix~\ref{reward:gamma}. With this annotation, each step $\tau_i$ receives a low-level toolcall reward $\gamma_i \in \{-1, 0\}$. Here, we deliberately set $\gamma$ to a penalty-only reward that punishes violations of the step without granting any positive reward, so the policy cannot exploit $\gamma$ itself to overwhelm other positive rewards. 
Consequently, we obtain a symbolic guard that largely prevents most commonly observed reward-hacking strategies against $\beta$.

\subsection{Lean4AgenticRL Training}\label{meth:train}

With the multi-level rewards in place, we unify the $\beta$ and $\gamma$ rewards with the verifier-based outcome reward $\alpha$ with the enhanced Lean4Agent formal framework that jointly manages all the rewards in a single reward server for scalable agentic RL. 
We apply this reward to optimize the policy with our variant of GRPO~\citep{guo2025deepseek}, namely \textit{workflow-stage-normalized GRPO}. Hereafter, we index the rollouts by $i \in [1, G]$, where $G$ is the group size in GRPO, the general workflow stages by $j \in [1, m]$, and steps within a stage by $k \in [1, n_{i, j}]$. Therefore, for each $\tau_{i, j, k}$, we define the reward schema as follows:
\begin{equation}
    r_{i, j, k}^{(0)} = \alpha_i + (0.2 + 0.3 \cdot \mathbb{I}\{\alpha_i = 1\}) \cdot \beta_{i, j}
\end{equation}
\begin{equation}
    r_{i, j, k}^{(1)} = \mathbb{I}\{\gamma_{i, j, k} = -1\} \cdot \min\{-1, r_{i, j, k}^{(0)}\} + \mathbb{I}\{\gamma_{i, j, k} = 0\} \cdot r_{i, j, k}^{(0)}
\end{equation}
\begin{equation}
    r_{i, j, k} = \mathbb{I}\{p \in \{0\%\} \cup [75\%, 100\%]\} \cdot r_{i, j, k}^{(1)} + \mathbb{I}\{p \in (0\%,  75\%)\} \cdot \alpha_i
\end{equation}
where $\alpha_i \in \{-1, 1\}$ is the high-level outcome rewards for rollout $i$, $\beta_{i, j}$ denotes the mid-level workflow stage reward, $\gamma_{i, j, k}$ is the low-level toolcall rewards, and $p$ is the group pass rate. 
In particular, to prevent the model from hacking the process rewards, we introduce a scale $\beta$ of at most 0.5 and apply it only to groups with an accuracy of 0\% or $\geqslant75\%$. 
Those are the groups that receive less stable gradients from outcome rewards, whereas our method can provide richer process rewards. For planning-phase tokens, we use the pure outcome reward $\alpha_i$ during GRPO training.

For the execution phase, we train the agent with the workflow-stage-normalized GRPO algorithm. Standard GRPO computes a trajectory-level advantage by assigning identical credit to each rollout step. For step-level process rewards, prior work~\citep{wang2026openclaw} applies direct alignments of tool calls, which compares steps that may serve unrelated purposes. In our algorithm, we leverage the segmented trajectory to compute the relative advantage among steps that are aligned to the same general workflow stage across different rollouts. Formally, the objective function for training the policy agent LLM $\pi$ with parameter $\theta$ is as follows:
\begin{equation}
\begin{aligned}
    \mathcal{J}_{GRPO} (\theta) 
    &= \mathbb{E}_{\left[\mathcal{P} \sim \mathcal{D}, \;\{\bm{\tau}^{(e)}_i\}_{i = 1}^G \sim \pi_{old}(\mathcal{E, W_S^{(i)}}, \tau_i^{(p)}) \right]} \frac{1}{G} \sum_{i = 1}^G \frac{1}{m} \sum_{j = 1}^{m} \frac{1}{n_{i, j}}\sum_{k = 1}^{n_{i, j}} \\
    &  \left( \min \left\{ \delta_{i, j, k} \cdot A_{i, j, k}, \text{clip} \left\{ \delta_{i, j, k}, 1 + \varepsilon_u, 1 - \varepsilon_l \right\} \cdot A_{i, j, k}  \right\} - \eta \cdot K_{i, j, k} \right)
\end{aligned}
\end{equation}
\begin{equation}
    \delta_{i, j, k} = \frac{\pi \left(\tau_{i, j, k} | \mathcal{P}, \bm{\tau}_{i}[:j - 1, :k - 1]\right)}{\pi_{old} \left(\tau_{i, j, k} | \mathcal{P}, \bm{\tau}_{i}[:j - 1, :k - 1]\right)}; \; K_{i, j, k} = \mathbb{D}_{KL} (\pi \| \pi_{ref})
\end{equation}
where $\pi_{ref}$ is the reference policy, $\varepsilon_{u/l}$ and $\eta$ are hyper-parameters, $\mathcal{P} = \langle\mathcal{E, S}\rangle$ is the problem draws from the training set $\mathcal{D}$, and $A_{i, j, k}$ is the workflow-stage and length normalized advantage calculated as follows:
\begin{equation}
    A_{i, j, k} = \frac{r_{i, j, k} - \text{mean}(\bm{G}_j)}{\text{std}(\bm{G}_j)}; \, \bm{G}_j = \cup_{i = 1}^G \left\{r_{i, j, q}\right\}_{q = 1}^{n_{i, j}}
\end{equation}
In particular, we normalize across the stages of the general workflow to ensure comparability between trajectories.
Through this normalization, we aim to overcome the step alignment problem in agentic RL by grouping steps around more meaningful workflow stages rather than low-level tool calls. Through the above steps, \method applies multi-level rewards to jointly make the training more effective.

\subsection{Summary}\label{meth:summary}

To provide a clear overview of the \method training framework, we present the complete framework as pseudocode in Appendix~\ref{appendix:alg}. We apply the \method training to train a family model named \textbf{LAR-9B/27B} (\textit{Agentic RL Model with Process Reward with Lean 4}) on the software-engineering (SWE) and general terminal-usage tasks.

%% file: sec/4_experiment.tex
\section{Experiment}\label{sec:exp}

We validate the effectiveness of \method and the performance of \textbf{LAR-9B/27B} model based on SWE-Bench-Verified~\citep{jimenez2024swebench}, TerminalBench-2.0 and 2.1~\citep{tbench_2025}. Specifically, we present the main results in Section~\ref{exp:results} and compare the training dynamics with outcome-reward GRPO in Section~\ref{exp:grpo}. Subsequently, we conduct ablation studies in Section~\ref{exp:abl} and case study in Section~\ref{exp:case}. We also perform additional experiments in Appendix~\ref{appendix:add_exp}.

\subsection{Experiment Setup}\label{exp:setup}

\subsubsection{Task and Benchmark}\label{setup:data}

We assess the \textbf{LAR} models in two advanced agentic tasks, namely, software engineering (SWE) and general terminal usage. They are measured by SWE-Bench-Verified~\citep{jimenez2024swebench} and TerminalBench-2.0 and 2.1~\citep{tbench_2025}, both of which are widely adopted in evaluating frontier LLMs~\citep{qwen3.5, zeng2026glm, team2026kimi3, anthropic2026opus46, xu2026deepseek, thinkingmachines2026inkling}. We evaluate on SWE-Bench-Verified under mean@3, and on TerminalBench-2.0 and 2.1 under mean@5, using randomized seeds. Further details on the benchmarks and tasks are presented in Appendix~\ref{appendix:bench_details}.

\subsubsection{Baselines}\label{setup:baseline}

To evaluate the effectiveness of the \method and the \textbf{LAR-9B/27B} models, we compare them with the latest open-weight models of similar size. Following previous works~\citep{ivison2026tmax, qwen3.5}, for the 9B model, we mainly compared with OpenThinker-Agent-V1~\citep{openthoughts-agent}, Qwen-3.5-9B~\citep{qwen3.5}, TMax-9B~\citep{ivison2026tmax}, and Gemma-4-12B~\citep{farabet2026gemma4}. The 27B-level model is compared with Qwen-3.6-27B~\citep{qwen3.6-27b}, TMax-27B~\citep{ivison2026tmax}, Qwen-3-Coder-30B~\citep{qwen3} Gemma-4-31B~\citep{farabet2026gemma4}, and Qwen-3.8-27B~\citep{qwen38}. All the baselines run with the highest reasoning effort and the longest supported context length to elicit their best performance. For the agent harness, we report the results on Terminus-2~\citep{Harbor_Framework} for TerminalBench-2.0 and 2.1 and Claude-code~\citep{anthropic2026claudecode} for SWE-Bench-Verified. 

\subsection{Implementation Details}\label{setup:implementation}

We train the \textbf{LAR-9B/27B} based on Qwen-3.5-9B~\citep{qwen3.5} and Qwen-3.8-27B~\citep{qwen38} using a mixture of TMax-15k~\citep{ivison2026tmax} and SWE-ReBench~\citep{badertdinov2026swe} dataset. Training is performed with asynchronous RL built on SLIME~\citep{slime_github} and OpenClaw-RL~\citep{wang2026openclaw}. The 9B model is trained for 160 steps, and the 27B model for 120 steps. Both with a batch size of 32 and group size of 32 $(32\times32)$ following GLM-5~\citep{zeng2026glm}. Follwoing the clip-higher strategy~\citep{yu2026dapo}, we set $\varepsilon_u = 0.28$ and $\varepsilon_l = 0.2$ and we use a learning rate of $1 \times 10^{-6}$ for both models. The maximum context length is set to 128K tokens with each response capped at 8K tokens. The training is conducted on a 64-H200 GPU cluster and takes 160 and 240 hours for the 9B and 27B models, respectively. Following Lean4Agent, we apply \texttt{v4.20.0} for Lean and Mathlib.

\begin{table*}
    \centering
    \vspace{-0.15in}
    \caption{Main experimental results of \textbf{LAR-9B/27B} compared to recent works using SWE-Bench-Verified, TerminalBench(TB)-2.0, and TB-2.1. The best performance models are \textbf{bolded} and the second-best performance models of a similar size are \underline{underlined}.}
    \label{tab:main}
    \vspace{0.05in}
    \small
    \begin{tabular}{ccccc}
        \toprule
        \textbf{Models}                                             & \textbf{Size}     & \textbf{SWE-Bench-Verified}     & \textbf{TB-2.0}     & \textbf{TB-2.1}  \\
        \midrule
        \textbf{OpenThinker-Agent-V1}~\citep{openthoughts-agent}     & 8B                & 15.7\%                & 1.9\%                 & -    \\
        \textbf{Qwen-3.5-9B}~\citep{qwen3.5}                         & 9B                & 44.0\%                & 19.3\%                & 16.1\% \\
        \textbf{TMax-9B}~\citep{ivison2026tmax}                      & 9B                & \underline{53.5\%}    & \underline{27.2\%}    & \underline{27.6\%}    \\
        \textbf{Gemma-4-12B}~\citep{farabet2026gemma4}               & 12B               & 23.2\%                & 15.3\%               & 27.0\%      \\
        \textbf{Qwen-3.6-27B}~\citep{qwen3.6-27b}                    & 27B               & 70.0\%                & 45.6\%                & 40.5\%        \\
        \textbf{TMax-27B}~\citep{ivison2026tmax}                     & 27B               & 70.0\%                & 49.9\%                & 44.9\%    \\
        \textbf{Qwen-3-Coder-30B}~\citep{qwen3}                      & 30B               & 49.2\%                & 24.5\%                & -         \\
        \textbf{Gemma-4-31B}~\citep{farabet2026gemma4}               & 31B               & 61.0\%                & 33.9\%                & 43.0\%    \\
        \textbf{Qwen-3.8-27B}~\citep{qwen38}                         & 27B               & \underline{83.0\%}    & \underline{64.7\%}    & \underline{69.6\%}   \\
        \midrule
        \textit{Our models} \\
        \midrule
        \textbf{LAR-9B}                                             & 9B                & \textbf{64.2\%}       & \textbf{32.8\%}       & \textbf{29.5\%} \\
        \textbf{LAR-27B}                                            & 27B               & \textbf{84.2\%}       & \textbf{68.2\%}       & \textbf{77.3\%} \\
        \bottomrule
    \end{tabular}
    \vspace{-0.15in}
\end{table*}

\begin{figure*}[t]
    \centering
    \begin{minipage}[t]{0.49\textwidth}
        \centering
        \includegraphics[width=\linewidth]{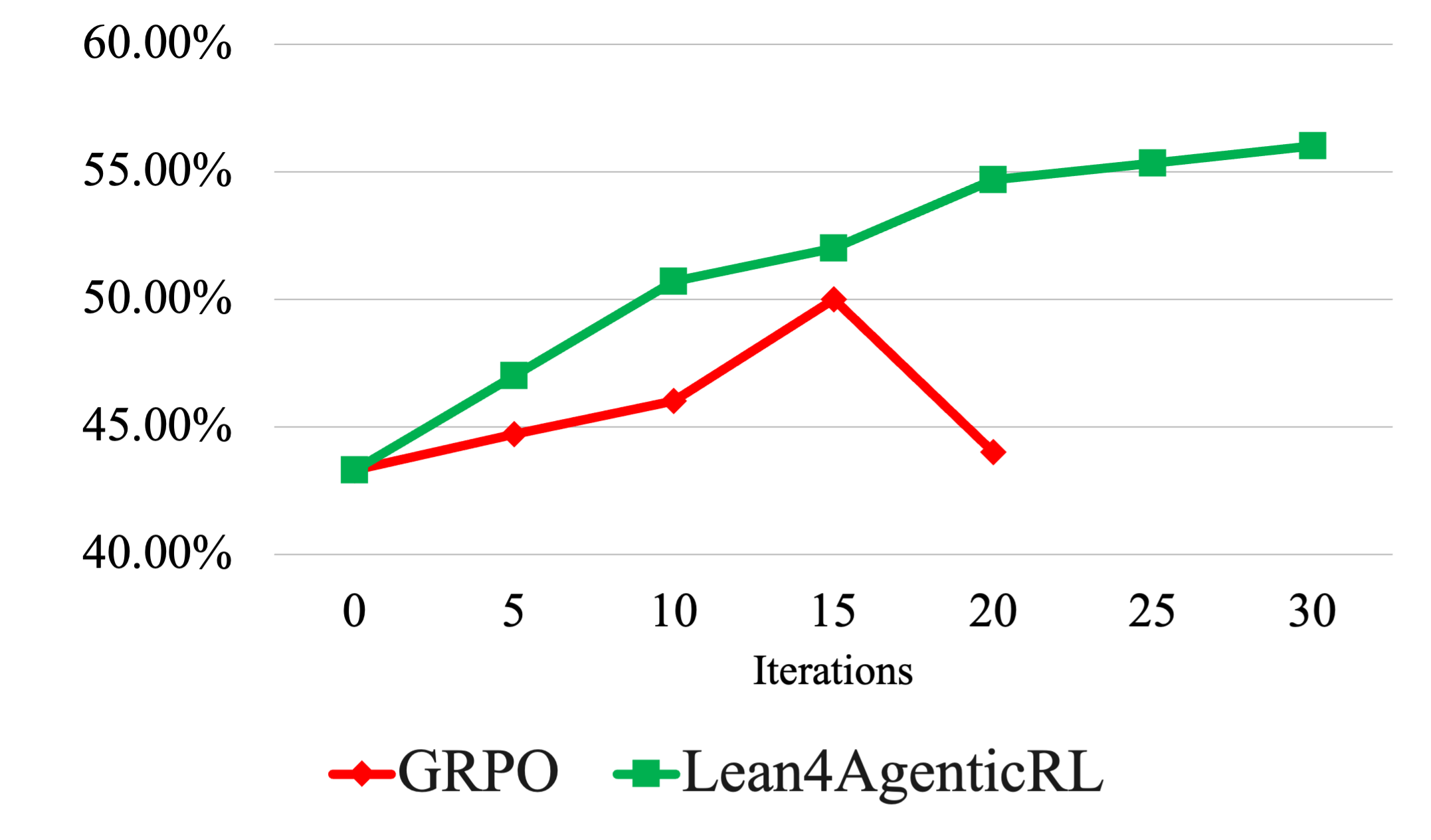}
    \end{minipage}\hfill
    \begin{minipage}[t]{0.49\textwidth}
        \centering
        \includegraphics[width=\linewidth]{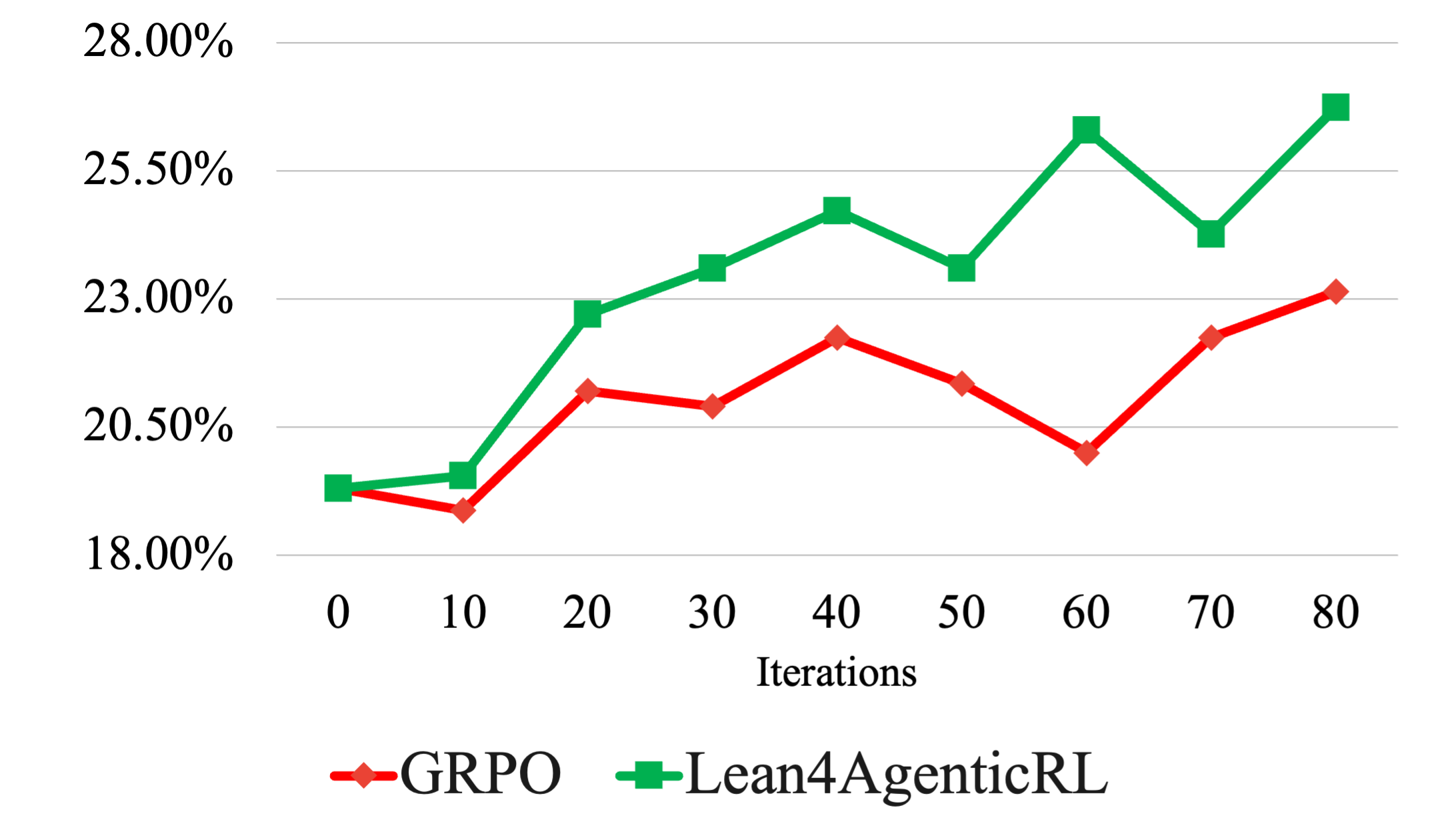}
    \end{minipage}
    \vspace{-0.15in}
    \caption{\method compared with GRPO in SWE by SWE-Bench-Verified and terminal usage tasks by TerminalBench-2.0.}\label{fig:rl_compare}
    \vspace{-0.2in}
\end{figure*}

\subsection{Main Results}\label{exp:results}

Table~\ref{tab:main} presents the results of \textbf{LAR-9B/27B} compared against open-weights models of similar size. At 9B scale, \textbf{LAR-9B} improves on its base model by \textbf{20.2\%}, \textbf{13.5\%}, and \textbf{13.4\%} on SWE-Bench-Verified, TerminalBench-2.0 and 2.1 with paired-p-value of average improvement$<10^{-10}$. It also surpasses the second-best model of comparable size by \textbf{10.7\%}, \textbf{5.6\%}, and \textbf{1.9\%} on those benchmarks with $p=3.5\times10^{-3}$. Notably, \textbf{LAR-9B} even exceeds the much larger Gemma-4-31B on SWE-Bench-Verified and performs on par with it on TerminalBench-2.0. At 27B scale, \textbf{LAR-27B} further raises the baseline's performance by \textbf{1.2\%}, \textbf{3.5\%}, and \textbf{7.7\%} on the three benchmarks respectively with $p=0.044$, reaching \textbf{84.2\%} on SWE-Bench-Verified and \textbf{77.3\%} on TerminalBench-2.1\footnote{At 27B scale, SWE-Bench-Verified is approaching saturation and TerminalBench-2.0 consists tasks whose external dependencies are inaccessible~\citep{terminalbench2026terminalbench21}, which makes TB-2.1 offering a more reliable comparison at this scale.}. 

In general, \textbf{LAR} family models achieve the best results at both scales across all three benchmarks, significantly outperforming their base models by \textbf{9.89\%} and the second-best model by \textbf{5.07\%} on average. These gains indicate that \method can provide more effective agentic RL that continuously optimizes models that have already undergone extensive post-training. We attribute this to the multi-level reward design, which obtains training signals from groups that originally produce less desirable gradients, thereby yielding more informative training signals. 

\subsection{Compared with Direct RL Training}\label{exp:grpo}

In this section, we investigate the performance of \method compared to normal GRPO-based agentic RL using pure outcome rewards under the OpenClaw-RL~\citep{wang2026openclaw} framework.
In particular, we conducted the training with a $16\times16$ step size using SWE-ReBench~\citep{badertdinov2026swe} for 30 steps on the SWE task, or TMax~\citep{ivison2026tmax} for 80 steps on the terminal-usage task, both based on Qwen-3.5-9B. Throughout the training, we evaluate the intermediate checkpoint on SWE-Bench-Verified with pass@1 and on TerminalBench-2.0 with mean@3.

The training dynamics are presented in Figure~\ref{fig:rl_compare}. The results show that \method outperforms the GRPO baseline throughout training on both tasks. On the SWE task, GRPO improves during the first 15 steps but then quickly deteriorates, falling back to roughly its initial level by step 20, while \method improves steadily over all 30 steps. 
We hypothesize that pure outcome rewards may reward some intermediate steps that are not functioning as desired, thereby destabilizing training at a later stage.
In the terminal usage task, GRPO fluctuates substantially and improves slowly, with a pronounced mid-task degradation, whereas \method maintains a clear upward trend with a consistent margin. Moreover, \method is more efficient in terms of training steps; it surpasses the best GRPO checkpoint within 10 steps on SWE and 30 steps on terminal-usage. Based on our full training, the performance to increase on both tasks as the training proceeds.

\subsection{Ablation Study}\label{exp:abl}

\subsubsection{Dropping Problem-Specific Workflow}\label{abl:wf}

We first examine whether the PSWF system itself benefits the agent execution without training. Specifically, we evaluate Qwen-3.5-9B on SWE-Bench-Verified in the Claude-code harness, with or without the PSWF. The results are shown in Table~\ref{tab:abl}, which shows that removing PSWF reduces the resolution rate from 44.0\% to 42.4\%. When we take a closer look at the failed trajectories, we observe that without PSWF guidance, the agent often explores the repository insufficiently, fails to locate the relevant files, and misattributes the root cause of the issue. This experiment demonstrates that, rather than limiting the agent's capabilities, applying the PSWF alone can improve the performance of LLM agents. 

\subsubsection{Dropping $\gamma$ rewards}\label{abl:gamma}

\begin{wraptable}{r}{0.35\textwidth} 
  \centering
  \vspace{-0.2in}
  \caption{Ablation Studies Result}
  \label{tab:abl}
  \vspace{0.05in}
  \small
  \begin{tabular}{cc}
    \toprule
    \textbf{Method}                 & \textbf{SWE-Bench-Verified} \\
    \midrule
    \textbf{w/o $\gamma$}           & 49.4\% \\
    \textbf{w/ $\gamma$}            & \textbf{56.0\%} \\
    \textbf{w/o PSWF}               & 42.4\%   \\
    \textbf{w/ PSWF}                & \textbf{44.0\%}   \\
    \bottomrule
  \end{tabular}
  \vspace{-0.2in}
\end{wraptable}

This experiment aims to assess the role of $\gamma$ by repeating the 30-step SWE training in Section~\ref{exp:grpo} with $\gamma$ removed from the reward, while keeping all other settings unchanged. As shown in Table~\ref{tab:abl}, removing $\gamma$ reduces the performance of the model by 6.6\%. It also increases the number of repeated commands captured by $\gamma$ by 66.8\% compared to the base model, while the full \method training observed almost no such behavior. Removing $\gamma$ also increases the total number of tool calls by 18.7\% compared to the full \method. Without this guard, the policy quickly learns to hack the LLM-as-judge predicates in $\beta$ through ritualistic actions, such as \texttt{echo} and repeated read-only or placeholder commands. As the training proceeds, these actions fill the context and inflate the response length with little contribution to the solutions. 
In contrast, with $\gamma$ in place, the response length of \method grows steadily alongside its performance, and we observed almost no reward hacking activities guarded by $\gamma$. These results confirm the necessity of the low-level toolcall rewards $\gamma$.

\subsection{Case Study}\label{exp:case}

This section illustrates the signal that our multi-level rewards provide during training. We evaluate the base model and the step-40, 80, and 120 checkpoints of \textbf{LAR-27B} on TerminalBench-2.1~\citep{tbench_2025} and annotate their trajectories using the training reward pipeline, as if they were rollouts in a group that receives process rewards. 
In particular, we select the \texttt{regex-chess} results at those checkpoints as the case we demonstrate. The detailed case is demonstrated in Appendix~\ref{appendix:case}.
As shown in Table~\ref{tab:case_summary}, the binary outcome reward $\alpha$ stays at $-1$ for the first three checkpoints and cannot distinguish them. In contrast, the fraction of stages with $\beta=+1$ rises from 2/8 to 4/7 and 5/6 before the task is solved at step 120 (5/5), while the model also writes fewer stages as training proceeds. When taking a closer look at the trajectories in Figures~\ref {fig:case_rc_0}, \ref{fig:case_rc_40}, \ref{fig:case_rc_80}, and \ref{fig:case_rc_120}, we can see that the workflow-stage reward accurately identifies the intermediate failures and successes.
This case illustrates that $\beta$ can provide a denser signal than $\alpha$ before a task is solved, and to some extent demonstrates the mechanism of \method's multi-level reward through detailed cases.

%% file: sec/2_related_work.tex
\section{Related Work}\label{sec:relat}

\subsection{Agentic Reinforcement Learning}\label{relat:rl}

Reinforcement Learning with Verifiable Rewards (RLVR) has become a scalable approach to enhancing LLM capabilities. It was first scaled by OpenAI-O1~\citep{openai2024reasoning} and open-sourced by DeepSeek-R1~\citep{guo2025deepseek} and Kimi-K1.5~\citep{team2025kimiK1.5}. RLVR shows that simple verifier-based outcome rewards can elicit reasoning capabilities without learning reward models. Building on this paradigm, agentic RL extends it to multi-turn agent interactions, typically rewarding the correctness of the final answer or task success~\citep{jin2025searchr1, zheng2025deepresearcher, wei2025webagent}. However, as the agent's task extends over a longer horizon, sparse outcome rewards that assign identical credit to every step cannot distinguish useful intermediate actions from redundant or erroneous ones. 
To provide denser training signals, subsequent studies apply process rewards on individual tool calls or infer step-level rewards from LLM judges~\citep{yu2025steptool, qian2025toolrl, liu2026agentic, xi2026agentprm, wang2026openclaw}. However, these signals are largely tied to low-level tool calls, which overlook broader progress and may even impair optimization~\citep{modecrua2026multi}. Prior work on hierarchical credit assignment~\citep{kazemnejad2024vineppo, feng2026group} relies on highly similar trajectories or tree-search methods, a setting that differs substantially from open-ended tasks such as SWE and general terminal usage.
A separate line of work replaces LLM judges with symbolic or algorithmic oracles as verifiable process rewards, but those oracles act as task-specific verifiers in controlled environments~\citep{yuan2026vpr, wang2025ma, wang2025gar}. In contrast, \method derives verifiable rewards across multiple levels under a unified formal language framework for general, open-ended agent executions. 

\subsection{Verifiable Agent Systems}\label{relat:agent}

As agentic systems mature and are increasingly deployed in high-stakes, long-horizon tasks, a growing line of work aims to formally model and verify agent behaviors. Some frameworks provide a type-based linter for the agent workflow that is embedded in agent languages~\citep{langchain2022, wang2026agentspex}, while others verify tool calls and their execution results~\citep{miculicich2025veriguard, liu2026toolgate, doshi2026towards}. More recently, Lean4Agent~\citep{wang2026lean4agent} formalizes agent behavior from workflow to execution in Lean, but it relies on predefined, fixed workflow settings. In contrast to theorem proving~\citep{wang2024theoremllama, ren2025deepseek, lin2025goedel} and code verification~\citep{li2026goedel}, where formal verifiers routinely supply training signals, these agent verifications serve only as runtime checks or post-hoc audits.

%% file: sec/5_conclusion.tex
\section{Conclusion}\label{sec:conc}

This paper presents \method, to the best of our knowledge, the first framework that derives multi-level process rewards under a unified framework based on Lean4, an expressive dependent-type formal language, for open-ended agentic RL. Its problem-specific workflow agent system balances the agent's inherent capability with execution controllability. \method proposes the workflow formalization methods that extend agent verification beyond fixed workflows to general harnesses. Built on this foundation, \method unifies high-level outcome rewards ($\alpha$), mid-level workflow-stage rewards ($\beta$), and low-level toolcall rewards ($\gamma$). It also proposes the workflow-stage-normalized GRPO algorithm, which aims to resolve the grouping problem in agentic RL. We train the \textbf{LAR-9B/27B} models using \method. Extensive experiments on SWE-Bench-Verified, TerminalBench-2.0, and 2.1 indicate that they outperform their base models by an average of \textbf{9.89\%} and the second-best model we know on those domains by \textbf{5.07\%}. To the best of our knowledge, \textbf{LAR-9B/27B} achieves the best results among models with comparable size. More broadly, \method lays a foundation for integrating verifiable symbolic signals into LLM training and points towards a new paradigm for agentic RL.

%% file: sec/6_others.tex

%% file: sec/X_Appendix.tex
\clearpage
\appendix

\etocdepthtag.toc{appendix}
\etocsettagdepth{mainbody}{none}
\etocsettagdepth{appendix}{subsection}
\etocsettocstyle{\section*{Appendix Contents}\vspace{-0.5em}\noindent\rule{\linewidth}{0.4pt}\par}{\noindent\rule{\linewidth}{0.4pt}\par}
\tableofcontents
\clearpage

\section{Limitations}\label{appendix:limitation}

Despite the strong results achieved by \method, our framework still has some limitations. Firstly, \method's verifiability still relies on explicitly bounding agent behavior externally; it does not open the black box of LLM's internal computation to achieve thorough and transparent verification.
Secondly, the scope of this work is limited to designing better rewards for agentic RL, without integrating the verification into the harness system at runtime.
Finally, \method primarily focuses on traditional agentic tasks, such as SWE or general terminal usage, without incorporating visual signals into the RL process.

\section{Term Definitions}\label{appendix:term}

Since the verifiable agent framework through dependent-type language like Lean is a new domain that may not be widely known by the community, we present the definition of the terms in this section:
\begin{enumerate}
    \item \textbf{Lean 4:} Lean 4 is a functional programming language and interactive theorem prover based on dependent type theory, whose trusted kernel mechanically checks the correctness of programs and proofs; in \method, leveraging the foundation library established in Lean4Agent~\ref{formal:lean4agent}, we use it as the shared formal foundation for building formal representations for workflows and unifying the multi-level reward signals.
    \item \textbf{Dependent type language:} A dependent type language, such as Lean, allows types to depend on values, so that a single language can express both programs and precise mathematical statements about them, and, under the Curry-Howard correspondence~\citep{sorensen2006lectures}, verifying a proof reduces to type-checking a term against the type of the statement it proves.
    \item \textbf{Agent workflow:} A structured plan for solving a task, represented as a directed graph $\mathcal{G} = \langle V, E \rangle$ whose nodes $V$ are workflow stages and whose edges $E \subseteq V \times V$ specify the allowed transitions between them, including sequential, branching, and loop-back transitions. In \method, we also refer to the natural-language PSWF, which carries this structure implicitly, as a workflow.
    \item \textbf{Workflow stage:} A node $v_i \in V$ of the workflow graph that represents a sub-goal of the task (e.g., reproducing the issue or verifying the fix) is called a workflow stage.
    \item \textbf{Agent step:} A single reasoning-action cycle of the agent, consisting of its reasoning, the toolcalls, and the observation of results, which serves as the atomic unit of a trajectory and the granularity at which the $\gamma$ rewards are assigned. Typically, a workflow stage consists of a sequence of agent steps.
    \item \textbf{Semantic predicates for agent:} Following Lean4Agent~\cite{wang2026lean4agent}, it is atomic decidable typed functions on the agent's execution state and trajectory evidence that describe atomic properties of agent behavior. The property can be verified by Lean or Python code when it is directly decidable, and by an evidence-grounded LLM judge when semantic interpretation is required.
    \item \textbf{Pre-/post-conditions:} Functions of predicates attached to each workflow stage $v_i$, where the pre-condition $\rho_i^{(\mathrm{pre})}$ specifies what must hold when the agent enters the stage and the post-condition $\rho_i^{(\mathrm{post})}$ specifies what the stage must establish upon exit, analogous to a Hoare triple $\{\rho_i^{(\mathrm{pre})}\}\ v_i\ \{\rho_i^{(\mathrm{post})}\}$.
\end{enumerate}

\section{Wrap up of \method}\label{appendix:alg}

We wrap up our \method training framework as the following algorithm:

\begin{algorithm}[H]
\caption{\method}\label{alg:GAR}
\begin{algorithmic}[1]
    \Require Problem set $\mathcal{D}$, base agent policy $\pi$, large-scale LLM $\pi^{(c)}$;
    \State $\pi, \pi_{old}, \pi_{ref} \leftarrow \pi$;       \Comment{Init the policies}
    \For{$t : [1, T]$} \Comment{Step $t$ of \method training}
        \State $\bm{\mathcal{P}} \sim \mathcal{D}$;      \Comment{Sample set of problem to resolve for this step}
        \For{$\mathcal{P} = \langle \mathcal{E, S} \rangle : \bm{\mathcal{P}}$}
            \State $\{\langle \mathcal{W_S}^{(i)}, \bm{\tau}_i^{(p)} \rangle \sim \pi(\mathcal{P, W})\}_{i = 1}^{G}$;       \Comment{Write PSWF in planning phase}
            \State $\{\langle \mathcal{E}'_i, \bm{\tau}_i^{(e)} \rangle \sim \pi(\mathcal{E}_i, \mathcal{W_S}^{(i)}, \bm{\tau}_i^{(p)})\}_{i = 1}^{G}$;      \Comment{Execute PSWF}
            \State $\{\mathcal{F}_{\mathcal{W_S}}^{(i)} \sim \pi^{(c)}(\mathcal{W}_\mathcal{S}^{(i)}, \mathcal{W})\}_{i = 1}^G$         \Comment{Formalize WF}
            \State $\{\{\tau_{i, j, k}\}_{j, k = 1}^{m_i, n_{i, j}} \leftarrow \text{Segment}(\pi^{(c)}, \bm{\tau_{i}^{(exe)}}, \mathcal{F}_{\mathcal{W_S}}^{(i)}) \}_{i = 1}^G$;        \Comment{Segment traj. according to WF}
            \State $\{\alpha_i \sim \text{Verifier}(\mathcal{E}_i')\}_{i = 1}^G$;        \Comment{Obtain outcome rewards}
            \State $\{\beta_{i, j} \leftarrow \beta\text{-annot}(\pi^{(C)}, \mathcal{F}_{\mathcal{W_S}}^{(i)}, \bm{\tau}_i^{(exe)})\}_{i, j = 1}^{G, m_i} $;       \Comment{Annotate WF with $\beta$ rewards}
            \State $\{\gamma_{i, j, k} \leftarrow \gamma\text{-annot}(\mathcal{F}_{\mathcal{W_S}}^{(i)}, \tau_{i, j, k})\}_{i, j, k = 1}^{G, m_i, n_{i, j}} $;       \Comment{Annotate WF with $\gamma$ rewards}
        \EndFor
        \State $\pi \leftarrow \text{Optimize}(\mathcal{J}_{GRPO}(\theta))$
    \EndFor
    \State \Return $\pi$
\end{algorithmic}
\end{algorithm}

\section{Additional Experiments}\label{appendix:add_exp}

\subsection{Applying \method training to additional models}\label{add_exp:more}

To demonstrate the effectiveness of \method, besides Qwen-3.5-9B~\citep{qwen3.5} and Qwen-3.8-27B~\citep{qwen38}, we also apply the training method to Qwen-3.6-27B~\citep{qwen3.6-27b} for 80 steps and DeepSeek-V4-Flash-0731-304B~\citep{xu2026deepseek} for 40 steps using TMax~\citep{ivison2026tmax} dataset following the exact same training recipes. 

For Qwen-3.6-27B base model, we evaluate the results using SWE-Bench-Verified~\citep{jimenez2024swebench} and TerminalBench-2.0~\citep{tbench_2025}. The base model's performance is 70.0\% and 45.6\%, respectively, while the trained model yields 74.0\% and 53.6\% on the results. This indicates an improvement of \textbf{4.0\%} on the SWE task and \textbf{8.0\%} on the terminal-usage task. For DeepSeek-V4-Flash-0731-304B, the base model achieves 67.3\% on TerminalBench-2.1, while our trained model achieves 72.9\%, representing a \textbf{5.6\%} improvement. These results indicate that the effectiveness of the \method training methods is consistent across model size, model family, and post-training variants. It further demonstrates the potential of further scaling the \method process reward to frontier models in significantly larger sizes.

\subsection{Faithfulness of Workflow Annotation}\label{add_exp:faith}

In this section, we evaluate the faithfulness of our LLM-based annotations during \method. In particular, we have the following three main parts of LLM-as-judge annotation in our framework:
\begin{itemize}
    \item \textbf{Predicate annotation faithfulness:} In this type of faithfulness, we evaluate whether the generated workflow in our problem-specific workflow (PSWF) stage has the correct workflow annotation. In particular, we evaluate whether it annotates correct pre- and post-conditions using the correct predicate. The annotation fails in this step if it misses some important predicates. For example, if the \texttt{acceptance\_criteria} predicate required by the PSWF is missing from the postcondition of the corresponding workflow stage, the annotation is counted as a failure.
    \item \textbf{Workflow segmentation faithfulness:} This type of faithfulness mainly verifies whether our segmentation process faithfully segmented the execution tool calls into the corresponding general workflow steps. The annotation fails at this step if the segmentation of tool calls spans significantly different workflow steps. For example, if one tool call in SWE agents is writing patches to fix the program, but it is classified as the \texttt{write\_test} workflow stage, then it is a failure.
    \item \textbf{Trajectory verification faithfulness:} It mainly covers the faithfulness of the process for our actual annotation of $\beta$ rewards with the trajectory. One annotation is counted as a failure if this workflow stage is judged a failure by a human/external LLM judge, but it provides $\beta$ rewards during training and vice versa. For example, if the \texttt{implement\_fix} workflow step fails to generate a correct fix for the SWE workflow, but in the training \method reward judgments provide $\beta$ reward to that step, then it is counted as a failure.
\end{itemize}

\begin{wraptable}{r}{0.4\textwidth} 
  \centering
  \vspace{-0.2in}
  \caption{LLM-based evaluation of workflow annotation faithfulness}
  \label{tab:faith_eval}
  \vspace{0.05in}
  \small
  \begin{tabular}{cc}
    \toprule
    \textbf{Faithfulness Type}                  & \textbf{Failure Rate} \\
    \midrule
    \textbf{Predicate annotation}               & 8.03\% \\
    \textbf{Workflow segmentation}              & 6.21\%\\
    \textbf{Trajectory verification}            & 4.84\%   \\
    \bottomrule
  \end{tabular}
  \vspace{-0.2in}
\end{wraptable}

In particular, the author team first manually evaluated a sample of 10 workflows to develop the judgment criteria for the three levels of faithfulness, and then applied Claude-code with Fable-5.1~\citep{anthropic2026claudefablemythos51} to annotate 1,200 trajectories from our training set. In total, for \textit{predicate annotation faithfulness}, we evaluated 32,295 predicate annotations, and for \textit{workflow segmentation faithfulness}, we evaluated 20,235 steps that segmented into 5,912 workflow stages, and for \textit{trajectory verification faithfulness}, we evaluated 5,912 $\beta$ reward grantings, matching the number of workflow stages we annotated. The failure rate is presented in Table~\ref{tab:faith_eval}.

The results show that the overall failure rate in our evaluation is below 10\%, which is relatively low when compared with previous efforts in formal reasoning LLM training~\citep{zhang2026physprover,lin2025goedel}
When taking a closer look at the detailed results, the failure cases show that most failures in predicate annotation occur in LLM-as-judge predicates, mainly due to ambiguity in the natural-language criteria relative to the workflow requirements. For workflow segmentation, most errors stem from ambiguity in segmentation. Due to the shared context across workflow stages and the relatively ambiguous indicators for each stage, the failure most commonly occurs in ambiguous cutting-point steps, such as the final verification of the last step, which can also be interpreted as a form of situation exploration for the next step. For trajectory verification, the most common failure pattern is a chunked context length of 16,384 tokens fed into the LLM judge, which, in some cases, results in context length overflow, causing the LLM judge to make judgments based on an incomplete context and leading to the failure. From the outstanding empirical results shown in Table~\ref{tab:main} and the relatively steady increase in Figure~\ref{fig:rl_compare}, there is no significant effect on training stability and performance at this error rate. A similar pattern was also observed on previous research on auto-formalization~\citep{wang2025kimina, lin2025goedel}.

To further examine the LLM-based faithfulness validator's results, we sampled 60 examples per annotation type, for a total of 180 faithfulness annotations. Among them, only one annotation in Predicate Annotation is judged as having an error. In particular, it misclassifies one predicate \texttt{coreCheckPatterns = "test"}, which is the checker for normalized bash commands. The LLM annotator treated \texttt{"test"} as test-time-only at the verification level, whereas in practice it is used to test the program in execution. Furthermore, providing this additional context to the annotator does not change its judgment of the predicate annotation, which the authors have subsequently verified to be correct. Thus, we conclude that our evaluation method for the three types of faithfulness of LLM-based workflow annotation is reliable enough to support the conclusions we draw from it.

\section{Comparison Between Dependent-Type with Other Formal Languages}\label{appendix:dt_to_other}

In this section, we compare the dependent-type language we apply in the Lean4Agent and the \method framework with other commonly used formal languages for the specification and verification of agent behaviors. Firstly, dependent-type languages like Lean offer greater expressiveness for uniformly modeling and verifying the agent workflows and trajectories. 
Other formal languages, such as propositional temporal logics like LTL, reason over a sequence of atomic propositions, which means they cannot describe structured data such as nested tool-call arguments or JSON schemas, cannot relate values across different states, and are limited to regular properties. Although it can express counting constraints over loops, more complicated cases, such as every edit being matched by a later test run, fall outside the logic.
Another commonly applied formal language is the Satisfiability Modulo Theories (SMT) language, such as Z3~\citep{de2008z3}. Although SMT solvers support recursive algebraic datatypes and recursive function definitions, their decision procedures are generally unable to prove the properties of self-referential definitions, since doing so requires induction, and some of the loop properties must be handled through induction.

In contrast, dependent type formal languages provide inductive types for recursive structures such as nested JSON and workflow graphs, support induction to reason properties about loops and self-referential definitions, and value-dependent types that allow a stage's pre- and post-conditions to be stated directly over the concrete data it consumes and produces, which is vital for self-verification behavior in agent executions. Furthermore, dependent type formal language also inherits the resolution of Russell's paradox~\citep{kneale1971russell} through higher-order statements that quantify over predicates and workflows while preventing the self-pointing constructions that render naive set theory inconsistent, whereas LTL and first-order SMT solver languages avoid such paradoxes only by being too weak to express higher-order statements at all. Moreover, temporal formulas and SMT-decidable fragments can be embedded as definitions within Lean, thus a single dependent-type formal language can host all of these formalisms while relying on a small trusted kernel to verify their results. These advantages are why we chose to use Lean as a uniform backbone for formalizing agent behaviors and providing multi-level rewards for agentic RL.

\section{Discussion of Formal Boundary for \method}\label{appendix:formal_boundary}

Since \method still applies LLM-as-judge in many places of the framework, we discuss the boundary between formal verification and informal model behavior in this section. Due to the black-box nature of LLM policy, we are unable to directly verify the interior generation process for the agents. Instead, following Lean4Agent~\citep{wang2026lean4agent}, we aim to bound the agent's behavior at each workflow stage by formulating pre- and post-conditions using the semantic predicate system. In particular, the formal boundary of the system is established after the problem-specific workflow has been formalized as Lean code, the execution trajectory has been segmented into workflow stages, and the annotation for LLM-as-judge predicates in the trajectory has been completed. These three steps are performed by LLMs and are unable to be formally guaranteed. Within the boundary, the Lean formulates the atomic predicates into a complex expression for modeling agent behaviors. Thus, the major source of informality in our system comes from errors in the LLM-as-judge annotation steps, which is detailed in Appendix~\ref{add_exp:faith}. 

We apply this set of formal boundaries to balance the expressiveness and formal guarantees of our reward system, with informality limited to the workflow formalization and the predicate judgment process. This provides more fine-grained control than simple rule-based LLM-as-judge verification and verifies more properties than traditional pure formal methods.

\section{Benchmark and Task details}\label{appendix:bench_details}

SWE-Bench-Verified contains 500 human-verified issues drawn from 12 widely used Python repositories on GitHub. Given an issue description and a containerized repository, the agent must produce a Git patch that resolves the issue, and its correctness is determined by repository-level test cases. In contrast, TerminalBench-2.0/2.1 comprises 89 challenging terminal-based tasks spanning SWE, system administration, security, scientific computing, and other general domains. The agent completes each task from its natural-language descriptions by executing commands in the container, and success is assessed by task-specific tests of the final container state. Following common practice, we report mean@3 on SWE-Bench-Verified and mean@5 on both TerminalBench versions, all under randomized seeds. We checked all three evaluation sets against our training data and found no overlap.

\section{LLM-based Annotation Details}\label{appendix:llm_annotation}

\subsection{Problem Specific Workflow Formalization}\label{llm_annotation:pswf}

This section provides a detailed explanation of how we formalize our problem-specific workflow as Lean code and verify it for Section~\ref{formal:lean4agent}. We use one of several frontier models (DeepSeek-V4-Pro, Claude-Opus-4.6/4.8, GLM-5.2, and Kimi-K3), depending on availability, with Claude-code to transform the SWE problem-specific workflow (PSWF) $\mathcal{W}_S$ and its corresponding task statements into a static semantic verification schema in the Lean4Agent~\citep{wang2026lean4agent} framework. The prompt for the formalization process is demonstrated in~Figure~\ref{fig:pswf-formalization}. We first construct the workflow schema, a graph whose nodes represent workflow stages and whose edges specify transitions, including retries and branches. 
Subsequently, we annotate the formal semantic representations by specifying the pre- and post-conditions for each stage, which specify the facts required before and after execution. It is supported by our enhanced predicate system, which is enriched in predicates in the SWE task and further extended to terminal-usage tasks.
The predicate here represents the atomic properties for agent behaviors.
In particular, the predicates represent variable-level properties of explicit variables in the workflows, information-level components that model the implicit information flowing in the process, and workflow-level predicates that model the high-level goals of the workflows.
The transformation of the PSWF into Lean4Agent verifiable form enables verification of internal consistency and lays the foundation for granting rewards in subsequent stages.

\subsection{Workflow Stage Segmentation}\label{llm_annotation:wf_seg}

This section details the process of our workflow segmentation. Its inputs are the PSWF, the formalized workflow, and the execution trajectory. The trajectory is a sequence of agent steps that include reasoning, tool calls, and their results. The agent step has no direct information about which workflow stage it belongs to. In this stage, we segment the trajectory into different stages. The details of the segmentation prompt are presented in Figure~\ref{fig:wfSeg}. 

\section{Reward Annotation Details}\label{appendix:reward}

\subsection{$\beta$ Reward Annotation}\label{reward:beta}
This section provides details of the method we used to annotate the $\beta$ reward. In particular, following Lean4Agent's design, the $\beta$ rewards determine whether the required postconditions are established by the stage's agent executions. For some predicates, we can verify them directly using Lean or Python code, while others may require LLM-as-judge verification. In particular, the latter handles properties that cannot be faithfully specified by programs. All the LLM judgments must be supported by observed evidence. The detailed annotation procedure is presented in Figure~\ref{fig:beta-skill}.

\subsection{$\gamma$ Reward Annotation}\label{reward:gamma}

In this section, we introduce the symbolic rules we judge for $\gamma$ rewards following Section~\ref{formal:tool}. We set the $\gamma$ reward as a negative-only guard over each agent step. The detailed rules are as follows:

\begin{enumerate}
    \item \textbf{Consecutive repetition:} This rule penalizes the third repeated back-to-back run of the same normalized bash command.
    \item \textbf{No-operation commands:} It penalizes the command chains consisting of no-operations or delays such as \texttt{exit, exit 0, true} or numeric \texttt{sleep x}.
    \item \textbf{Constant shell output:} It penalizes constant-only \texttt{echo}, \texttt{printf}, or print-only \texttt{cat + constant string} chains that neither inspect nor modify the task state.
    \item \textbf{General pure-print script:} It penalizes the Python or other form of code snippets whose syntax tree contains only literal-valued \texttt{print} calls and no state-dependent executions.
    \item \textbf{Intermediate no-action:} It penalizes the non-final agent step with no executable bash or tool actions to prevent the agent from inflating the step count with reasoning-only turns.
    \item \textbf{No-progress re-execution:} It penalizes re-running the same bash command within a workflow stage with no environment-modifying tool calls.
\end{enumerate}

During training, fast Lean + Python code evaluates the above rules on the trajectory and automatically passes them to the reward server for uniform reward assignment. 
We use Python tools to parse the initial string into syntax trees, and we represent and verify its properties through predicates in Lean.
In particular, we penalize only the steps, without rewarding the agent in $\gamma$, to avoid the reward itself becoming the source of hacking. In general, this set of rules can largely prevent the agent from hacking the rewards by inflating the number of steps the agent takes. Combined with the empirical study in Section~\ref{abl:gamma}, we conclude that the method relatively is effective in our experiment settings.

\newpage
\begin{figure}[!htbp]
\centering
\begin{minted}[breaklines, bgcolor=verylightgray, fontsize=\scriptsize]{markdown}
# SWE-Bench problem-specific workflow formalization

ROLE
Translate the PSWF into a Lean4Agent workflow contract.

INPUT
- PROBLEM: Problem statement
- SPECIFIED_WORKFLOW: Problem-specific workflow
- DOMAIN SCHEMA: generic SWE stages, Lean skeleton, and supported predicates

STRUCTURE
1. Assign distinct node ids and explicit entry/exit ids. Preserve each action's concrete instruction and typed reads/writes.
2. Construct a workflow graph schema for the PSWF. Encode transitions as seq, loop, loopBack, or branch edges. Introduce loops/branches only when the written plan describes them.
3. Annotate the predicate information of each node and combine it into variable-level (pre/postcondition chain), information-level (info_requires closure), and workflow-level (sub-goal and graph-predicate coverage) predicate status for the Lean verdicts.
4. For a loop, retain its invariant, termination specification, and exit postconditions; for a branch, identify both successor nodes.

DEPENDENCIES AND GOALS
1. Analyze the basic predicates precond_extra/postcond_extra using supported predicates. Declare parameter bindings and shell_run availability at entry.
2. Encode concrete fact dependencies in info_requires and their sources in produces_context_info / param_produces_variable_info. Preserve missing upstream support; do not invent an established fact.
3. Link goal_spec.sub_goals to evidence variables, contributing nodes, verifying nodes, and required graph predicates.

OUTPUT AND VALIDATION
Emit only the formalized Lean code based on the skeleton. Evaluate it in Lean variable/Hoare, information-flow, and coverage verdicts independently. Repair compilation errors only; a negative consistency verdict is valid. Retain the final generated Lean, verdict, and conversion provenance. Mark reconstruction or skeleton fallback as degraded.
\end{minted}
\caption{SWE-Bench workflow formalization prompt's general schema with detailed prompts/examples abridged due to space constraints.}
\label{fig:pswf-formalization}
\end{figure}

\begin{figure}[!htbp]
\centering
\begin{minted}[breaklines, bgcolor=verylightgray, fontsize=\scriptsize]{markdown}
# SWE-Bench workflow-stage segmentation

ROLE
Assign each trajectory step to a node in the workflow. Judge boundaries locally, not from the full rollout.

INPUT
- WORKFLOW: topologically ordered semantic nodes with intents/postconditions
- CURRENT_STEP, NEXT_STEP, and optional LOOP_BACK_TARGET
- STEP: reasoning + tool call(s) + short result slices

PREPROCESS
Coalesce a reasoning-only message with the following tool-bearing message. Keep a trailing narration-only message as the terminal turn.

FOR EACH STEP (cursor starts at the workflow entry)
1. Judge the action actually executed; narration about a future step is not evidence that the boundary has been crossed.
2. Return exactly one decision:
   STAY      - the step still advances CURRENT_STEP;
   ADVANCE   - the step starts NEXT_STEP and belongs to it;
   LOOP_BACK - only across a declared loop edge; start a new visit index.
3. Prefer STAY when ambiguous. Never make an undeclared backward jump.

SWE ACTION CUES
- Read/Grep/Glob or read-only Bash on source       -> explore_repository
- write/run a bug-specific reproduction           -> reproduce_issue
- Edit/Write a non-test source file                -> fix_issue
- run a relevant test or rerun the reproduction   -> verify_fix
- generate/read the final git diff or submit patch -> create_patch
Use the action's goal, not its tool name alone; Bash can belong to any stage.

OUTPUT
For every turn emit {turn_idx, step, menu_idx, visit, decision, tool_ids}. Require exactly one assignment per turn and preserve trajectory order.
\end{minted}
\caption{SWE-Bench workflow segmentation prompt's general schema with detailed prompts/examples abridged due to space constraints.}
\label{fig:wfSeg}
\end{figure}

\begin{figure}[!htbp]
\centering
\begin{minted}[breaklines, bgcolor=verylightgray, fontsize=\scriptsize]{markdown}
# Stage-level execution evaluation

PURPOSE
Assess whether the step in the trajectory meets the requirements of our formal workflow stages.

INPUT
- The task description and the stage's intended objective.
- The variable/information/workflow level predicates associated with the stage.
- The stage's assigned tool calls, observed outputs, and artifact evidence.

PROCEDURE
1. Locate evidence relevant to each predicate requirement. Distinguish the agent's claims and intentions from actions and observed results.
2. Check directly decidable properties against the recorded evidence directly through Lean verification. Use an LLM for requirements that need semantic interpretation.
3. Apply the relevant software-repair criterion:
   EXPLORATION  - code evidence supports the defect diagnosis;
   REPRODUCTION - the reproducer captures the reported failure;
   REPAIR       - the change addresses the cause of the defect;
   VERIFICATION - executed checks support the corrected behavior;
   SUBMISSION   - the submitted patch contains the intended changes.
4. For verification, inspect which tests ran and what they observed. A statement that tests passed is not evidence that they ran.
5. Record support, contradiction, or insufficient evidence, together with an explanation tied to the observed actions and results. Do not invent missing results or change the stage requirements.

OUTPUT AND REWARD
Combine the condition judgments into a stage-completion verdict. For the paper's binary reward rule:
  beta_i = +1 if the required postconditions are established;
  beta_i = -1 otherwise.
Attach this stage reward to the steps assigned to that stage.
\end{minted}
\caption{Workflow stage-level $\beta$ annotation prompt's abridged skeleton.}
\label{fig:beta-skill}
\end{figure}
\newpage
\section{Case Study on Learning Progression across \method Training}\label{appendix:case}

\subsection{Overall Analysis of the Case}\label{case:overall}

This section introduces the details for our case study in Section~\ref{exp:case}. We aim to study the training process in detail by analyzing the dynamics of evaluating intermediate checkpoints to elucidate the mechanism. Note that all evaluations are performed post-hoc and are not used to select the best model checkpoint. Such a demonstration intends to illustrate the performance of our training and explain why it works by presenting a case. In particular, we select the evaluation results of \texttt{regex-chess} for the model checkpoints at the 0-th (base model), 40-th, 80-th, and final 120-th steps, formalizing the PSWF, segmenting the execution-phase trajectory, and annotating the corresponding $\beta$ rewards. The overall reward status for the model training is presented in Table~\ref{tab:case_summary}.

\begin{table}[h]
    \centering
    \caption{Overall $\alpha$ and $\beta$ rewards change as model training progresses}
    \label{tab:case_summary}
    \vspace{0.05in}
    \setlength{\tabcolsep}{4pt}
    \begin{tabular}{cccc}
        \toprule
        \textbf{Step}   & \textbf{TB-2.1 Test cases}    & \textbf{$\alpha$}     & \textbf{Stages with $\beta=+1$ / total stages} \\
        \midrule
        0 (base)        & 1/4 passes                    & -1                    & 2/8 \\
        40              & 1/4 passes                    & -1                    & 4/7 \\
        80              & 2/4 passes                    & -1                    & 5/6 \\
        120             & 4/4 passes                    & +1                    & 5/5 \\
        \bottomrule
    \end{tabular}
\end{table}

From the table, we can see that as training progresses, although the model is unable to finish that problem, the $\beta$ accuracy keeps increasing, indicating that our \method reward design is prompting the model to solve additional problems throughout the process. And the model can finally resolve that problem as the process reward steadily grows. This trend indicates that \method has the potential to enable the model to acquire capabilities it originally lacks, which has proven to be an important topic in RLVR~\citep{chen2026does}. 

\subsection{Detailed Case}\label{case:detail}

This section presents the abridged execution-phase trajectories of the four checkpoints in Figures~\ref{fig:case_rc_0}, \ref{fig:case_rc_40}, \ref{fig:case_rc_80}, and \ref{fig:case_rc_120}; each segmented into PSWF stages with the corresponding $\beta$ rewards. The base model (Figure~\ref{fig:case_rc_0}) builds a correct FEN codec and a size-valid \texttt{re.json}, and both stages receive $\beta=+1$. However, a malformed lookahead in its move-rule builder means no move rule ever fires, so the move-rule, output-field, and final verification stages receive $\beta=-1$. At step 40 (Figure~\ref{fig:case_rc_40}), the model builds a move generator that covers castling and en passant, and four of its seven stages earn $\beta=+1$. Its remaining failures come from a king-safety filter whose knight and pawn offsets are off by one, and from a rule tail that does not accept the castling field. At step 80 (Figure~\ref{fig:case_rc_80}), the legality filter handles knight checks and pins, queen promotion is tested, and the positions that broke step 40 now pass, which raises the verifier result to 2/4 tests and gives $\beta=+1$ on five of six stages. Both remaining failures stem from castling out of check, which the castling guard misses because the PSWF item only requires that the path be empty and the rights be present. At step 120 (Figure~\ref{fig:case_rc_120}), the model correctly rejects castling out of check, every stage earns $\beta=+1$, and all four tests pass. Taken together, $\alpha$ stays at $-1$ for the first three checkpoints, while $\beta$ gives increasingly fine-grained credit for the stages each attempt actually establishes. This provides a learning signal before the task is solved, which benefits the learning process.

\begin{figure}[!htbp]
\centering
\begin{minted}[breaklines, bgcolor=verylightgray, fontsize=\scriptsize]{markdown}
# regex-chess | base model (step 0) | 20 exe-phase steps
# Outcome: alpha = -1 (verifier 1/4: only test_not_long)
# Task: write /app/re.json, a list of [regex, repl] pairs mapping a white-to-move FEN to
#       all legal next FENs (castling, en passant, Q-promotion); < 100k pairs, < 10 MB
# PSWF: 8 stages;

[PSWF_01] 4x4 'state' prelude + legality-mask table        steps 0-1    beta=-1 (2/3)
  t0  $ cat > /tmp/g.py <<'PY' ... PY          (64x64 legality-mask generator, chunk 1)
  t1  "My previous /tmp/g.py chunk was broken" -> $ rm -f /tmp/g.py   (design dropped)
[PSWF_03] Board normalisation (FEN decode/encode)          steps 2-4    beta=+1 (3/3)
  t2  $ python3 /tmp/test_decode.py -> PER-SQUARE: rnb.k.nr/p..p.ppp/...  len: 71
  t3  $ python3 /tmp/test_encode.py -> ROUNDTRIP OK: False   ('.' matched any char)
  t4  ('z' as empty) -> ENCODED: rnb1k1nr/p2p1ppp/3B4/1p1NPN1P/6P1/3P1Q2/P1P5/q4Kb1
      -> ROUNDTRIP OK: True                                        <- cited evidence
[PSWF_04] Move rules for every origin square               steps 5-9    beta=-1 (3/3)
  t5  $ python3 /tmp/parse_stage.py -> re.PatternError: bad escape \x at position 2
  t6  (fix) -> WORKING: 'rnbzkznr/.../qzzzzKbz\x010011\x02--\x03'  LEN: 80 (expect 80)
  t7  $ python3 /tmp/attack_test.py -> ATTACK TEST OK
  t8-9 $ cat > /tmp/gen.py ...; cat >> /tmp/gen.py ...   (move-rule builder, not run)
[PSWF_05] Ep-field restore; checkpoint "check.py -> OK"    step 10      beta=-1 (3/3)
  t10 $ python3 /tmp/gen.py -> pairs: 3907  bytes: 696266    (check.py is never run)
[PSWF_02] Turn-flip pairs; checkpoint "re.json < 100000 pairs, < 10MB"
                                                           step 11      beta=+1 (2/3)
  t11 $ python3 /tmp/apply.py "rnb1k1nr/... w kq - 0 1" -> LINES: 1  ''
      cited evidence: t10 "bytes: 696266" (3907 pairs: the size limits are met)
[PSWF_06] Output-field ('b'/'w') pairs                     steps 12-13  beta=-1 (3/3)
  t12 $ python3 /tmp/dbg.py -> 3873 80 -> 0 | ''   became empty at 3873
  t13 $ python3 /tmp/dbg2.py -> 26 no patlen 72  27 no patlen 71 ... 39 no patlen 110
      pattern: '^(?=.0][N])(?=.11][bknpqrz])(.{1})(.{9})...'  (no move rule matches)
[PSWF_07] Terminal / strip pairs                           steps 14-16  beta=-1 (3/3)
  t14 "Root cause: lit() produced (?=.0][N]) -- missing the {} braces"
  t15 $ python3 /tmp/fixlit.py -> NOT FOUND: "(=?=." + str(to_line(L)) + "][" ...
  t16 $ python3 /tmp/fixlit2.py -> a found: False c found: True
      -> return "(?=." + str(to_line(L)) + "}" + "[" + cls(cs) + "])"  (still broken)
[final_verification]                                       steps 17-19  beta=-1 (3/3)
  t17 NoOp (response had no valid JSON)
  t18 -> line16: return "(=?=." + "{" + str(to_line(L)) + "}" + ...    (typo remains)
  t19 prints line 16 again; re.json is never regenerated; the 7,200 s budget ends it
Verifier: start position -> "Our move:  not found" (re.json turns every FEN into '');
  the 3 game tests fail, test_not_long (size limit) passes. No gamma fires.
Stage rewards: PSWF_02/PSWF_03 steps -0.8, others -1.2
\end{minted}
\vspace{-0.1in}
\caption{The abridged execution-phase trajectory for the step 0 (base) model in \texttt{regex-chess}}
\label{fig:case_rc_0}
\end{figure}

\begin{figure}[!htbp]
\centering
\begin{minted}[breaklines, bgcolor=verylightgray, fontsize=\scriptsize]{markdown}
# regex-chess | step 40 | 95 exe-phase steps
# Outcome: alpha = -1 (verifier 1/4: only test_not_long)
# PSWF: 7 stages;

[PSWF_01] Reformat: digits -> '.', 71-char board + marker    steps 8-21   beta=+1 (3/3)
  t13 gen.py rule tail '( [wb] )(K?)(Q?)([kq]*)( [^@]*) @@@': no match if castling '-'
  t16 P/N/K/B/R/Q rule loops -> move rules: 2361;   t18-19 -> SAMPLE OUR: 0
  t20 -> after flatten: 'rnb.k.nr/p..p.ppp/...B..../.p.NPN.P/......P./...P.Q../
      P.P...../q....Kb. w kq - 0 .'                               <- cited evidence
[PSWF_02] Pseudo-legal moves via backreference duplication   steps 22-29  beta=-1 (2/3)
  t22-27 tail-group bugs ('\S' ate the '@' of the marker; '(.* )$') -> match? None
  t28 -> SAMPLE OUR: 43 ... '...6P1/3P2Q1/P1P5/q4Kb1 b kq -'        <- cited (rep1)
      = 43 of the 45 pseudo-legal moves (a2-a4, c2-c4 missing; t40: OURS 12 PY 20 on
      the start position); t29: in check, legal ['f1g2', 'f1e2', 'f3d1']; no ep yet
      rep2/3 judges also +1, but cite t50 (after the stage) -> from_the_future -> -1
[PSWF_05] King-safety filter (1st summary after t30)        steps 30-52  beta=+1 (3/3)
  t39-50 gen2.py rebuilt (t22-27 rule tail kept) -> start position: 20 lines, ep set
  t51 build_filter: 'p.{8}K', 'p.{10}K', 'n.{11}K(?=...)', 'n.{7}K(?=...)',
      'n.{19}K(?=...)', 'n.{17}K(?=...)', mirrored K..n terms, rook/bishop rays.
      With 9 chars per rank these gaps are 1 too large: knight/pawn checks are missed
  t52 sample -> COUNT 3: '...P.Q../P.P.K.../q.....b. b kq - 0 0' ... <- cited (rep1/2)
      (rep3: t52's Qd1 line; the check is along rank 1, no knight/pawn case is tested)
[PSWF_06] Assemble re.json; len < 100000, size < 10 MB       steps 53-56  beta=-1 (3/3)
  t53 -> rules 2401 size 414197 / ValueError: too many values to unpack
  t54-56 (R.append fix) -> wrote 2391 / OURS 3 PY 3 ... / OURS 20 PY 20 ...
      all 3 judges +1, all downgraded from_the_future: rep1/2 cite a t56 line that
      the judge's clipped view elides (found at t72, t93); rep3 cites t72
[PSWF_03] Castling: O-O / O-O-O, rights, rook move           steps 57-67  beta=+1 (2/3)
  t57 $ python3 check.py -> AssertionError: 44 != 45 ... RNB1K2R w KQkq - 2 7
  t61 "make_castle_OO/000 patterns have no (?s)" -> OURS 45 PY 45 countmatch True
  t66 O-O-O (.{3} for f1-h1) -> OURS 43 PY 43 countmatch True subset True <- cited (rep2)
  t67 $ python3 check.py -> Ran 1 test in 9.475s / OK                <- cited (rep3)
      rep1/2 judge t77-94 as PSWF_03: rep2 +1 cites t66; rep1's t50 cite is clipped -> -1
[PSWF_04] En passant (+ promo "handled in step2")        steps 0-7, 68-92  beta=+1 (3/3)
  t6  python-chess probe -> epsq 20 legal False token -   (ep shown only if legal)
  t68-69 ep-capture rules, king adjacency, castle-safety lookaheads -> filter2 26
  t73 check.py OK; '.../3pP3/... w KQkq d6 0 3' -> OURS 30 PY 31 (no exd6)
  t75-76 ('d6' is flattened to 'd......') epName fix -> OURS 31 PY 31 ... <- cited (3/3)
  t77-91 fuzz vs python-chess: TypeError, 2x AttributeError, killed twice: no result
  t92 '8/8/8/8/8/3kK3/8/8 w - - 0 1' -> OURS 1 PY 4, 'O:' (empty); promotion untested
[final_verification] Acceptance checklist                    steps 93-94  beta=-1 (2/3)
  t93 "The degenerate adjacent-kings FEN is illegal (out of scope, ignore)"; castle
      fix ' b  '->' b - ' -> wrote 2414 size 421707; 4=4 (in check), 12=12, 20, 3
  t94 fuzz restarted in background; check.py -> Ran 1 test in 9.540s OK <- cited (rep3)
      rep1/2 judged t57-67 as this stage: -1. The 2nd context summary then runs into
      the 20,000 s budget; the fuzz never reports
Verifier: test_not_long passes; each game passes 12-22 position checks (base: 0):
  Immortal, move 22 '.../q5b1 w - - 1 22': "Our move:  not found" (1 vs 39): input
    castling '-' fits no rule tail (t13 design, kept in gen2; same empty output at t92)
  Century, move 13: 42 vs 41 (Ke2 into the Nc3 knight); Naroditsky, move 12: 39 vs 3
    (the Nd3 check is ignored): both are the t51 knight offsets
Stage rewards: PSWF_01/03/04/05 steps -0.8, PSWF_02/06/final -1.2
\end{minted}
\vspace{-0.1in}
\caption{The abridged execution-phase trajectory for the step 40 model in \texttt{regex-chess}}
\label{fig:case_rc_40}
\end{figure}

\begin{figure}[!htbp]
\centering
\begin{minted}[breaklines, bgcolor=verylightgray, fontsize=\scriptsize]{markdown}
# regex-chess | step 80 | 121 exe-phase steps
# Outcome: alpha = -1 (verifier 2/4: test_game_of_century, test_naroditsky_ivanchuk fail)
# PSWF: 6 stages;

[PSWF_01] Re-confirm baseline (probe reprints the 3 lines)  steps 0-1    beta=-1 (3/3)
  t0  $ python3 -c " import chess ... n=b.push(m) f=n.fen().split() ... "
      -> AttributeError: 'NoneType' object has no attribute 'fen'
  t1  $ tail -n 50 /app/check.py   analysis: "Sample position has exactly 3 legal white
      moves and the set matches expected output"  (t0 printed only the traceback)
      rep2/3: judge -1; rep1: judge +1 citing t63 (quote not found) -> downgraded
[PSWF_02] Diff harness /tmp/harness.py      step 2 (rep2/3: 13-23)       beta=+1 (2/3)
  t2  $ cat > /tmp/gen_re.py <<'PYEOF' ... -> syntax ok    (rep1's segment; no harness)
  t13 (rep2/3 seg.) $ cat > /tmp/harness.py ... -> invalid group reference 3
  t23 -> ... our=2 exp=3 FAIL  <- cited (rep2/3; checkpoint wants count >= exp: 2 < 3)
      (rep1: the quote lies after its t2 segment -> downgraded "from the future")
[PSWF_03] Builder /tmp/gen_re.py -> /app/re.json  steps 3-5 (rep2/3: 2-12)  beta=+1 (3/3)
  t5  $ python3 /tmp/gen_re.py -> part A loading 171 (preproc + pawns)  <- cited (rep1)
  t11-12 (rep2/3 seg.) -> king 947 / rook 1899 / ... / castle 4031, then the legality
      filter (mark lines whose white K is attacked, delete them)
      -> markers 6341 / total 6352 / size 1001073  <- cited (rep2/3; crashes at t13)
[PSWF_04] Implement a-g (moves, captures, Q-promotion, ep, castling, legality, rights)
          and test each with the harness                     steps 6-63   beta=+1 (3/3)
  t33 -> rnb1k1nr/p2p1ppp/3B4/1p1NPN1P/6P1/3P1Q2/ our=3 exp=3 OK  <- cited (rep1/3)
  t35 $ timeout 600 python3 /app/check.py -> Our move: .../PPPPKPPP/RNBQ1BNR b KQkq -
      not found (Ke2 keeps KQ); t41-49 K/R rights stripping -> t53 Ran 1 test ... OK
  t59 -> 4k3/8/8/2pP4/1P6/8/8/4K3 w - c6 0 2  our=9 exp=9 OK  (en passant); mate and
      stalemate FENs our=1 exp=0 FAIL -> t60 "Only impossible 0-move cases fail now"
  t61 -> FUZZ: ALL OK positions: 42; t62 (Q-promotion) k7/2n1P3/... our=4 exp=4 OK
[PSWF_05] Keep size under budget (count pairs + bytes)       steps 64-70  beta=+1 (3/3)
  t65 raw sample vs its own re-ordered expected (Qd1 first; task: Ke2, Kg2, Qd1)
      -> line 0 same / line 1 same / line 2 same     (task order restored at t93)
  t69 -> Ran 1 test in 72.898s / OK / pairs: 14177 size: 2263441  <- cited (rep1/3)
[final_verification] check.py, exact sample, size, edge FENs steps 71-120 beta=+1 (3/3)
  t71 Complete -> (context summary) t72 "I need to re-verify everything myself"
  t77-86 finds + fixes 3 rights bugs (a8/h8 k/q swap, promotion captures keep rights,
      duplicate castling lines for rights K/KQ); t95-110 python-chess-style
      normalisation of the input castling field
  t111 -> ALL EDGE TESTS: PASS; t114 -> fuzz done: 50 white positions, fails=0
  t117 -> sample lines == expected: True count: 3 valid 2-item pairs: True
      pairs: 14249 size: 2283900                                  <- cited (rep1/3)
  (rep2 cites t111 for PSWF_04, t91 for PSWF_05, t116 "Ran 1 test" for this stage)
Verifier: test_immortal_game (24 positions) and test_not_long pass; the other two games
  fail first on castling out of check: after 16...Rfe8+ (.../3RK2R w K) "Our moves: 7,
  Python-chess moves: 6", after 17...Qa5+ 11 vs 10; both "Our move: .../3R1RK1 b - -
  not found" (O-O). Castling's guard notatt() (t6) sees a file/diagonal slider only if
  every char between it and e1 is '.', e.g. (?<![rq]\.{62}) (t38), but each such run
  crosses a '/', so no file or diagonal check is ever seen; the PSWF's castling item
  asked only "guard K/R path empty & rights present". No gamma fires.
Stage rewards: PSWF_01 steps -1.2, others -0.8
\end{minted}
\vspace{-0.1in}
\caption{The abridged execution-phase trajectory for the step 80 model in \texttt{regex-chess}}
\label{fig:case_rc_80}
\end{figure}

\begin{figure}[!htbp]
\centering
\begin{minted}[breaklines, bgcolor=verylightgray, fontsize=\scriptsize]{markdown}
# regex-chess | step 120 | 173 exe-phase steps
# Outcome: alpha = +1 (verifier 4/4)
# PSWF: 5 stages;

[PSWF_01] Write the builder that emits re.json; checkpoint: pair count and byte size
          printed, within limits                           steps 0-42   beta=+1 (2/3)
  t16 "compression used '.' (any char) instead of literal dot. Fixing to '\.'"
  t28 "The empty-square test used an unescaped '.' (any char)" -> re.escape('.')
  t39 -> count 3: '...P1P1K3/q5b1 b kq -' '...P1P3K1/q5b1 b kq -' '...q2Q1Kb1 b kq -'
  t42 $ python3 /app/gen.py -> pairs 9827 bytes 1663928
      fen legal 20 ours 20 MATCH / fen legal 3 ours 3 MATCH        <- cited evidence
[PSWF_02] Output fields: side b, castling subset, ep only if capturable, ' 0 0'
                                                           steps 43-45  beta=+1 (3/3)
  t44-45 refactor to make_line/edit -> pairs 9827 ... fen legal 3 ours 3 MATCH
      (cited: t39 sample line 'rnb1k1nr/.../P1P1K3/q5b1 b kq - 0 0')
[PSWF_03] Validate incrementally: sample, check.py, fuzzer; report PASS/FAIL counts
                                                           steps 46-86  beta=+1 (3/3)
  t47 castling O-O/O-O-O, ep capture, 4-variant double push added
  t70 sample (3) and start position (20) both MATCH
  t86 $ python3 /app/t.py -> DIFF legal 31 ours 30 '.../3pP3/... w KQkq d6 0 1'
      ... DONE bad= 2 / 23                                         <- cited evidence
[PSWF_04] Iterate until the fuzzer is clean; re-check size steps 87-137 beta=+1 (3/3)
  t96 "file = s%9 (s%8 is off). This corrupted all (file,rank) conversions"
  t118 board-only digit expansion [str(d)+'(?=.* w)','.'*d]  (ep 'd6' keeps its 6)
  t128 -> EXIT=0  DONE bad= 0 / 23                                 <- cited evidence
  t130 empln.py (check.py's split) -> total elements: 36  empty at indices: [35]
  t135 final rule r'\n+$' -> total elements: 35  empty at indices: []
[final_verification]                                      steps 138-172 beta=+1 (3/3)
  t139 -> Successfully tested 18 positions from Morphy's Opera Game (33 moves)
                                                                   <- cited evidence
  t163 -> FUZZ DONE passed 70 failed 1 of 71   (the 1: k7/8/8/8/8/8/8/R5K1 w, black
       already in check; the oracle "captures" the king)
  t169 -> 4400579 /app/re.json  pairs 17437 ok True / SET MATCH: True count 3 3
Verifier: all 4 tests pass; castling out of check is now refused (Century, move 17:
  6 vs 6). No gamma fires.
Stage rewards: every step 1.5
\end{minted}
\vspace{-0.1in}
\caption{The abridged execution-phase trajectory for the step 120 model in \texttt{regex-chess}}
\label{fig:case_rc_120}
\end{figure}